\documentclass{article}

\usepackage[final, nonatbib]{bdl_2018}

\usepackage[utf8]{inputenc} % allow utf-8 input
\usepackage[T1]{fontenc}    % use 8-bit T1 fonts
\usepackage{hyperref}       % hyperlinks
\usepackage{url}            % simple URL typesetting
\usepackage{booktabs}       % professional-quality tables
\usepackage{amsfonts}       % blackboard math symbols
\usepackage{nicefrac}       % compact symbols for 1/2, etc.
\usepackage{microtype}      % microtypography
\usepackage{graphicx}
\usepackage{amsmath}
\usepackage{mathtools}

\newcommand{\Vol}{\mathrm{Vol}}
\newcommand{\KL}{\mathrm{KL}}
\newcommand{\MM}{\mathcal{M}}
\newcommand{\DD}{\mathcal{D}}
\newcommand{\LL}{\mathcal{L}}
\newcommand{\QQ}{\mathcal{Q}}
\newcommand{\EE}{\mathbb{E}}
\newcommand{\RR}{\mathbf{R}}

\title{Embedding-reparameterization procedure for manifold-valued latent variables in generative models}

\author{
  Eugene Golikov \\
  Neural Networks and Deep Learning Lab \\
  Moscow Institute of Physics and Technology \\
  Russia \\
  {\tt golikov.ea@mipt.ru}   \\\And
  Maksim Kretov \\
  Neural Networks and Deep Learning Lab \\
  Moscow Institute of Physics and Technology \\
  Russia \\
  {\tt kretov.mk@mipt.ru} 
}

\begin{document}

\maketitle

\begin{abstract}
  Conventional prior for Variational Auto-Encoder (VAE) is a Gaussian distribution. Recent works demonstrated that choice of prior distribution affects learning capacity of VAE models. We propose a general technique (embedding-reparameterization procedure, or ER) for introducing arbitrary manifold-valued variables in VAE model. We compare our technique with a conventional VAE on a toy benchmark problem. This is work in progress.
\end{abstract} 

\section{Introduction}

Variational Auto-Encoder (VAE) \cite{vae} and Generative Adversarial Networks (GAN)~\cite{gan-orig} show good performance in modelling real-world data such as images well. The key idea of both frameworks is to map a simple distribution (typically Gaussian) of lower dimension to a high-dimensional observation space by a complex non-linear function (typically neural network). Most of research efforts are concentrated on the enhancement of training procedure and neural architectures giving rise to a variety of elegant extensions for VAE and GANs~\cite{gan-overview}.

We consider prior distribution that is mapped to data distribution $p(x)$ as one of design choices when building generative model. Its importance is highlighted in a number of works~\cite{elbo-surgery, s-vae18, homeo-vae, sphere-nlp}. Although~\cite{s-vae18} provides an extensive overview of usage of $L_2$-normalized latent variables (points lying on a hypersphere); this is clearly just one of the possible design choices for prior distribution in generative model.

Recent works~\cite{s-vae18, homeo-vae, sphere-nlp} argued that manifold hypothesis for data~\cite{belkin} provides evidence in favor of using more complicated priors than Gaussian, for which the topology of latent space matches that of the data. The above mentioned works derived analytic formulas for reparameterization of probability density on manifold (hypersphere in~\cite{s-vae18} and Lie group $SO(3)$ in~\cite{homeo-vae}).

\begin{figure}[h]
  \centering
  \includegraphics[scale=0.18]{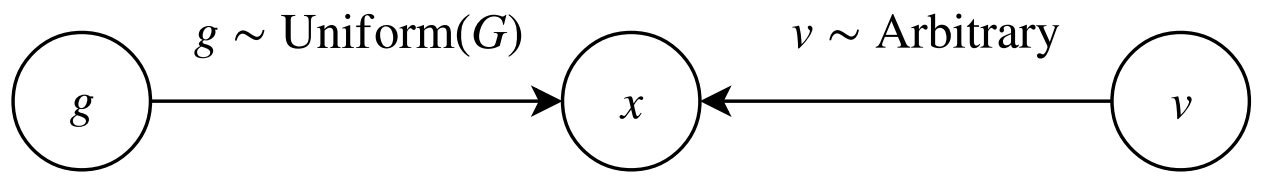}
  \caption{Observed data $X$ are generated by \textit{uniform} sampling from compact symmetry group $G$ and other independent factors of variation $V$ (for example, label of the class).}
  \label{sym} 
\end{figure}

A somewhat less rigorous argument in favor of using manifold-valued latent variables is that we can represent generative process for data as having two sources of variation (see Figure~\ref{sym}): one is uniform sampling from a group of transformations that we consider as compact symmetry groups (for example group of rotations) and another one is all the rest. This favors the choice of such topology of the latent space that would match "real" generative process: choose uniform distribution on some compact symmetry group as a prior distribution for latent variables.

Once a universal procedure for fast prototyping of VAE with different manifold-valued variables is available, such VAE can be used for estimating the likelihood integral $p(x|Model)$ (for example using IWAE estimate~\cite{iwae}) and thus make conclusions about latent symmetries that are present in the data. This was one of the key motivations for the current work.

All of above brings to the focus the case of continuously differentiable symmetry groups (Lie groups), which is a special case of manifold-valued latent variables. 

\section{Manifold-valued latent variables}

Let us make the following preliminary assumption:

\textit{Data $x \sim p(x), x \in S \subset \mathbf{R}^n$ are generated as on Figure~\ref{sym} with Lie group $G$ embedded in $\mathbf{R}^m$ and there is a continuous mapping $G \rightarrow S$.}

When using images as a test bed it implies that images generated by "close" symmetry elements (say two similar rotation angles $\phi_1$ and $\phi_2$) are also close in the pixel space. It justifies using additional tricks such as continuity loss~\cite{homeo-vae} for training VAE with manifold-valued latent variables.

%Seemingly trivial, because it is hard to invent counter example, this presumption may not hold even for toy datasets: consider images of thin line rotated by angles from 0 to 360 degrees. If there are not much samples, two close angles may have just as much overlap as two distant angles. So if this assumption does not hold true, task seem quite desperate: we do not have any hints how ordering in the group $G$ works.

\subsection{Construction of VAE}

Recall the optimization problem for VAE~\cite{vae}:
$$
\LL(\phi, \psi) = 
\EE_{x \sim \DD} \left[\EE_{z \sim q_\phi(z | x)}[\log p_\psi(x | z)] - \KL (q_\phi(z | x) \| p(z))\right] \to \max_{\phi, \psi},
$$
where $\DD$ denotes the data distribution, $q_\phi(z|x)$ is a posterior distribution on latent space $Z$, $p(z)$ is the corresponding prior, and $p_\psi(x|z)$ is the likelihood of a data point $x$ given $z$.
In order to construct a VAE with manifold-valued latent variables, we need the following:
\begin{enumerate}
    \item An encoder that produces the posterior distribution $q(z|x)$ from a parametric family of distributions on a manifold.
    \item An ability to sample from this posterior distribution: $z \sim q(z | x)$.
    \item An ability to compute KL-divergence between this posterior and a given prior.
\end{enumerate}

Recent works~\cite{s-vae18, homeo-vae} proposed approaches to working with manifold-valued latent variables that are similar in spirit to ours: they derive a reparameterization of probability density defined on smooth manifold and use it in VAE. Problem is that such derivation appears to be complicated and needs to be done for all manifolds of interest.

Our approach is the following.
First of all, we introduce a hidden latent space $Z_{hid}$, such that $\mathrm{dim} \, Z_{hid} = \mathrm{dim} \, \MM = n$, where $\MM$ is our manifold lying in a latent space $Z$ of dimension $m > n$. 
Let $p(z_{hid})$ be a prior distribution on $Z_{hid}$.

Suppose then, we have an embedding $f:\; Z_{hid} \to Z$, so that $f(Z_{hid}) \subset \MM$. Being an embedding requires $f$ to be a diffeomorphism with its image, in particular, $f$ should be a differentiable injective map. We also pose an additional constraint on $f$: it should map the prior on $Z_{hid}$ to a prior on the manifold $\MM$; in other words, if $z_{hid} \sim p(z_{hid})$, then $f(z_{hid}) \sim p(z)$.

Using this embedding $f$, we can construct a VAE with manifold-valued latent variables as depicted on the right part of Figure~\ref{fig:wae_and_manifold_latent_vae}.
In this case the posterior distribution $q(z_{hid}|x)$ on $Z_{hid}$ together with the embedding $f$ induce a posterior distribution $q(z|x)$ on $\MM \subset Z$. 
We then have to compute KL-divergence between this induced posterior and the prior $p(z)$ on the manifold.
Despite the fact that in this case the probability mass is concentrated on the manifold $\MM$ and hence the probability density on $Z$ is degenerate, we can define the manifold probability densities $q_{\MM}(z|x)$ and $p_{\MM}(z)$ (see Appendix 5.1 for details). Moreover, the corresponding KL-divergence is equivalent to the KL-divergence between distributions defined on $Z_{hid}$ (Appendix 4.3):
$$
    \KL(q_{\MM}(z|x) \| p_{\MM}(z)) =
    \KL(q(z_{hid}|x) \| p(z_{hid}))
$$

Hence the final optimization problem for model on the right part of Figure~\ref{fig:wae_and_manifold_latent_vae} becomes the following:
$$
\LL(\phi, \psi) = 
\EE_{x \sim \DD} \left[\EE_{z_{hid} \sim q_\phi(z_{hid} | x)}[\log p_\psi(x | f(z_{hid}))] - \KL (q_\phi(z_{hid} | x) \| p(z_{hid}))\right] \to \max_{\phi,\psi},
$$
where $\phi$ are parameters of VAE encoder, which encodes the object $x$ into $Z_{hid}$ space, and $\psi$ are parameters of VAE decoder which maps the manifold $\MM \subset Z$ to data-manifold in feature space; $\DD$ is our data distribution.

Thereby working with probability distributions induced on manifold of interest is easy: both terms in VAE loss (reconstruction error and KL-divergence) are easily calculated in the original hidden space $Z_{hid}$ that is further mapped on a manifold. 
%We leave the investigation of limitations and expressivity of induced probability distributions in described model remains future work.

\subsection{Learning manifold embedding}

\begin{figure}[h]
    \centering
    \includegraphics[width=0.8\textwidth]{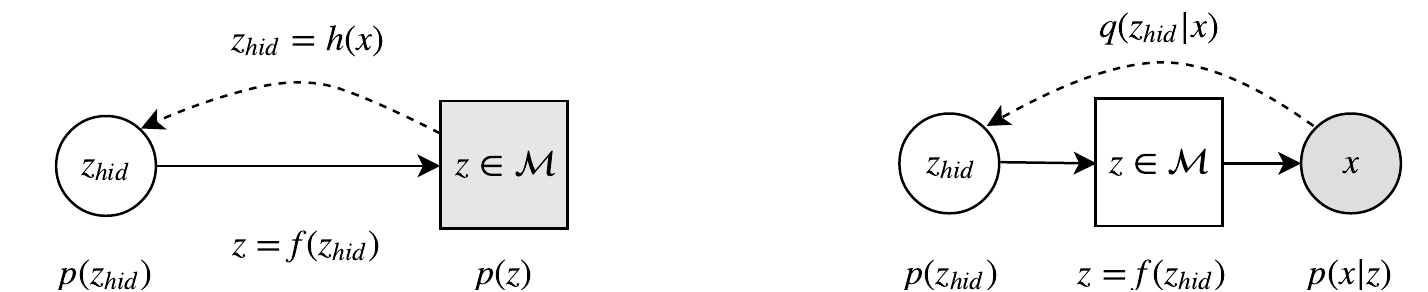}
    \caption{\textbf{Left:} Scheme diagram of WAE with feature space $Z$ and latent space $Z_{hid}$ that learns the prior distribution $p(z)$ on manifold $\MM \subset Z$. \textbf{Right:} The generative model with manifold-valued latent variables $z$. The squared node is deterministic.}
    \label{fig:wae_and_manifold_latent_vae}
\end{figure}

To apply the procedure described above, we have to construct an embedding $f$.
In order to do this, we propose the following procedure:
\begin{enumerate}
    \item Sample data from $p(z)$ (distribution on $\mathcal{M}$).
    \item Train Wasserstein Auto-Encoder (WAE)~\cite{wae} on the data from $p(z)$ (feature space) and the latent space $Z_{hid}$ with the prior $p(z_{hid})$: see the left part of Figure~\ref{fig:wae_and_manifold_latent_vae}.
    \item Use the decoder of this trained WAE as our embedding function $f$.
\end{enumerate}
Our motivation is the following: since the dimension of latent space $Z_{hid}$ and the dimension of manifold $\MM$ are the same, the reconstruction term in WAE objective constraints its decoder to be an injective map. Since it is represented with a neural network, it is also differentiable. The objective of WAE learning also forces its decoder to map a prior distribution on a latent space (in our case, $p(z_{hid})$) to a distribution of data to the feature space (in our case, $p(z)$). Hence WAE decoder is an ideal candidate for an embedding $f$.

%We propose to use different approach, which we called the embedding-reparameterization procedure. This approach is two-step. 
%\begin{enumerate}
%    \item "Embedding step". Learn the auxiliary generative model for manifold of interest. This model is further used for sampling and estimating the probability density on that manifold. Wasserstein Auto-Encoder (WAE, MMD variant)~\cite{wae} with uniform prior for its latent variables $z_{hid}$ (which we refer as "hidden latent variables") was applied for this role (Figure~\ref{fig:wae_and_manifold_latent_vae}, left). We selected this type of generative model because it allows for deterministic mapping from hidden latent space $Z_{hid}$ to its feature space $Z$.
%    \item "Reparameterization step". Decoder of this auxiliary generative model is used for reparameterization of manifold-valued latent variables $z$ (Figure~\ref{fig:wae_and_manifold_latent_vae}, right). 
%\end{enumerate} 

%In order to learn mapping $\mathbf{R}^n \supset Z_{hid} \rightarrow \mathcal{M} \subset Z = \mathbf{R}^m$, we use WAE with feature space $Z = \mathbf{R}^m$ and latent space $Z_{hid} = [0,1]^n$. Let $p(z_{hid})$ be some prior on latent space (i.e. uniform on $[0,1]^n$), and $p(z)$ be a prior on a manifold $\MM$. Objective of WAE learning requires its decoder $f$ to injectively map the prior in latent space to a prior on a manifold: i.e., if $z_{hid} \sim p(z_{hid})$, then it should be $f(z_{hid}) \sim p(z)$. The decoder can be made deterministic.
%It is then further included in VAE as a part of the decoder (square deterministic node on Figure~\ref{fig:wae_and_manifold_latent_vae}).

At first glance the described model leaves quite similar questions as vanilla VAE: we "shifted" the complex task of learning non-homeomorphic manifolds of a different topology (latent space and data space) from the VAE decoder to sub-module of the same VAE but pretrained using WAE. Nevertheless, the procedure ensures better control over mapping to manifold and one can develop corresponding metrics to control the quality of mapping.

\section{Introducing symmetries of latent manifold into encoder}

Recall that in our scheme an encoder $q(z_{hid}|x)$ together with embedding $f: \; Z_{hid} \to Z$ induce a family of posterior distributions $q(z|x)$ on $\MM$; let us call this family $\QQ$.

A natural requirement to $\QQ$ is to have the same symmetries as $\MM$ has. Suppose we have a symmetry group $G$ of $\MM$ acting on $Z$, i.e.
$$
    \forall z \in \MM, \, \forall g \in G \quad gz \in \MM.
$$
For example, if $\MM$ is an $n$-dimensional sphere $S^n$ in $Z = \RR^{n+1}$, $G$ is a group of rotations $SO(n+1)$.
We require $G$ to also be a symmetry of $\QQ$ also:
$$
    \forall q \in \QQ, \, \forall g \in G \quad \exists q' \in \QQ: \;  \forall z \in \MM \quad q'(gz) = q(z).
$$
This means that if a symmetry $g$ of $\MM$ acts on samples $z$ from a distribution $q \in \QQ$, we should get samples from another distribution $q'$ from the same family $\QQ$. 
Note that we did not pose this requirement while training $f$, hence it would not generally be satisfied. Therefore we have to symmetrize $\QQ$ explicitly.

\begin{figure}
    \centering
    \includegraphics[width=0.5\textwidth]{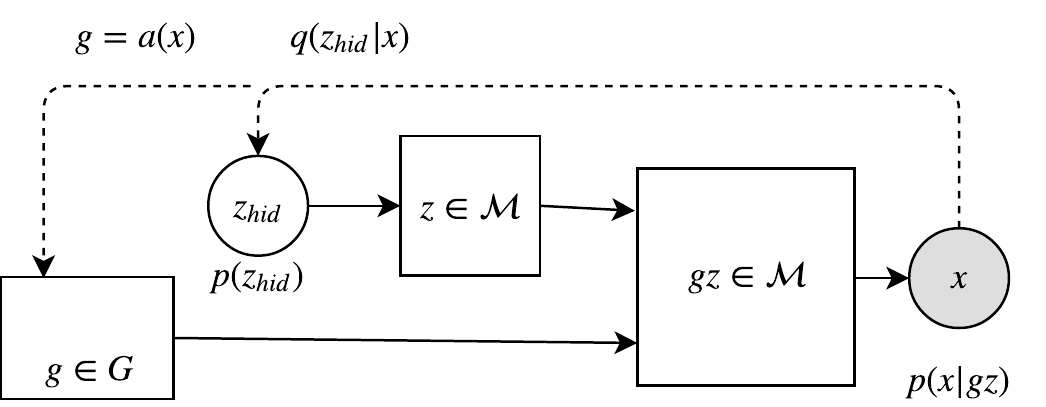}
    \caption{The generative model with group action-encoder acting on manifold-valued latent variable $z$.}
    \label{fig:manifold_latent_det_action_vae}
\end{figure}

In order to do this we introduce a group action encoder $a(x)$, see Figure~\ref{fig:manifold_latent_det_action_vae}. This group action encoder produces an element $g = a(x)$ of the symmetry group $G$ of $\MM$, which further acts on a sample $z = f(z_{hid})$. This effectively enriches the posterior family  $\QQ$ with $q': \; q'(gz) = q(z)$.

This procedure has close connection with homeomorphic VAE~\cite{homeo-vae}. Suppose our manifold $\MM$ is a compact Lie group. Then it is homeomorphic to its own symmetry group: $\MM \cong G$. Then our group action-encoder is equivalent to $R_\mu$ of~\cite{homeo-vae}.

\section{Experiments and conclusions}

\begin{table}
  \caption{Results on the toy task for different models.}
  \label{results}
  \centering
  \begin{tabular}{lc}
    \toprule
    Models & ELBO \\
    %\cmidrule(r){2-3}  \cmidrule(r){4-5} \\
    \midrule
    %Beta-Gauss, 1d & $175.3 \pm 3.4$ \\
    %Beta-WAE(MMD)-Gauss, 1d & $203.0 \pm 29.8$ \\
    %Beta-Projection-Gauss, 1d & $184.3\pm 10.9$ \\
    %Beta-WAE(MMD)-Action-Gauss, 1d & $\mathbf{247.4 \pm 45.9}$ \\
    %Beta-Gauss, 2d & $\mathbf{330.0 \pm 31.2}$ \\
    VAE, $\mathrm{dim}\, Z = 1$ & $183.98 \pm 11.66$ \\
    Manifold-latent VAE with learned $f$, $\mathrm{dim}\, Z_{hid} = 1$ & $197.19 \pm 20.46$ \\
    Manifold-latent VAE with $f = f_{proj}$, $\mathrm{dim}\, Z_{hid} = 1$ & $193.40 \pm 24.57$ \\
    Manifold-latent VAE with learned $f$ and group action encoder, $\mathrm{dim}\, Z_{hid} = 1$ & $\mathbf{259.03 \pm 59.14}$ \\
    VAE, $\mathrm{dim}\, Z = 2$ & $\mathbf{356.53 \pm 22.96}$ \\
    \bottomrule
  \end{tabular}
\end{table}

We followed the same experimental setup as for a toy task in paper~\cite{s-vae18}, but without noise. \footnote{Our code is available on GitHub: \url{https://github.com/varenick/manifold_latent_vae}} Sampling of a batch from the dataset consisted of two steps:

\begin{enumerate}
    \item We generated uniformly distributed points on a 1-dimensional unit sphere embedded in $\textbf{R}^2$.
    \item We applied a non-linear fixed transformation $\RR^2 \rightarrow \RR^{100}$ implemented as a randomly initialized multilayer perceptron with one hidden layer of size 100 and ReLU nonlinearity. Xavier-uniform initialization scheme was applied to the hidden layer. %Gaussian noise (0.1 standard deviation and zero mean) was added to the final transformation.
    %Two options were considered: with addition of Gaussian noise (0.1 standard deviation and zero mean) and without one.
\end{enumerate}

All models are VAEs with the posterior distribution $q(z_{hid}|x)$ (Beta on $[0,1]^n$), the prior distribution $p(z)$ (uniform of $[0,1]^n$) and the likelihood $p(x|z)$ (Gaussian on $\RR^{100}$). As for the reparameterization function $f(z_{hid})$, it was either WAE-MMD or the exact mapping from segment $[0,1]$ into a 1-dimensional circle ("Projection") in the first layer of decoder:
$$
    f_{proj}(z_{hid}) = \begin{pmatrix}
    \cos(2 \pi z_{hid}) & \sin(2 \pi z_{hid}))
    \end{pmatrix}^T.
$$
The dimensions of latent variables $n$ were either 1 or 2. 

In a case when the group action encoder is used, it produces an angle (element of $SO(2)$), which is further used to rotate the sample $z = f(z_{hid}) \in \RR^2$.

%We have also considered a special case ("Action"), when the encoder of VAE model produces not only the parameters of posterior distribution, but also an angle (group action). The sample of this posterior distribution $z_{hid}$ is then fed into decoder of pretrained WAE (reparameterization function $f(z_{hid})$) to get a sample $z$ from some distribution on a 1-dim sphere. We then used the produced angle to rotate this sample on a sphere. This rotated sample is then fed next into decoder of VAE. See Appendix 4.4 for details and motivation for this case.

The results are presented in Table~\ref{results}. All decoder structures that include manifold mapping show better results than a vanilla VAE with 1-dimensional latent Gaussian space.

\subsubsection*{Acknowledgments}

This work was supported by National Technology Initiative and PAO Sberbank project ID 0000000007417F630002.

\bibliographystyle{unsrt}
%\bibliography{nips_2018}

\section{Appendix}

\subsection{Probability density functions with manifold support}

Suppose we have a probability distribution on $Z=\mathbf{R}^n$ with density $p(z)$ and a diffeomorphism $f: Z \to X$, where $X = \mathbf{R}^n$ as well. Then, $f$ induces a probability distribution on $X$ with the following density:
$$
p(x) = 
p(f^{-1}(x)) |\det J_f(f^{-1}(x))|^{-1} =
p(f^{-1}(x)) |\det J_{f^{-1}}(x)|.
$$

Suppose now that $X = R^m$ with $m > n$, and $f: Z \to X$ is a smooth embedding (which requires $f$ to be a diffeomorphism between $Z$ and $f(Z)$). From this follows that $f$ induces degenerate probability distribution on $X$ since all the probability mass in $X$ is concentrated on a manifold $\mathcal{M} = f(Z)$. The corresponding probability measure is trivial:
$$
P(f(A)) = P(A)
$$
for some event $A$ on $Z$. Although we cannot define a valid probability density of $X$, we can define a manifold probability density on $\mathcal{M} = f(Z)$ as follows:
\begin{equation*}
    \begin{split}
        p_\mathcal{M}(f(z)) &\coloneqq
        \lim_{\Vol(A) \to 0 \; s.t. \; z \in A} \frac{P(f(A))}{\Vol_\mathcal{M}(f(A))} \\&=
        \lim_{\Vol(A) \to 0 \; s.t. \; z \in A} \frac{P(A)}{\Vol_\mathcal{M}(f(A))} \\&=
        \lim_{\Vol(A) \to 0 \; s.t. \; z \in A} \frac{\int_A p(z) \, dz_1 \ldots dz_n}{\Vol_\mathcal{M}(f(A))} \\&=
        p(z) \lim_{\Vol(A) \to 0 \; s.t. \; z \in A} \frac{\int_A dz_1 \ldots dz_n}{\Vol_\mathcal{M}(f(A))},
    \end{split}
\end{equation*}

where by $\Vol_\mathcal{M}(f(A))$ we denote an $n$-dimensional volume of $f(A) \subset \mathcal{M}$; let us define this volume. Let $\Omega$ be an open subset of $Z$. Then its image under embedding $f$ is an open subset of a manifold $f(\Omega)$ (open in terms of the topology of $\mathcal{M}$). If $Z$ is a Euclidean space, than the "volume" of $\Omega$ is given simply as:
$$
\Vol(\Omega) = \int_{\Omega} dz_1 \ldots dz_n.
$$
Since $\mathcal{M}$ is embedded into $X$, and $X$ is a Euclidean space, we can measure an $n$-dimensional "volume" of $f(\Omega) \subset \mathcal{M}$. It is given as:
$$
\Vol_\mathcal{M}(f(\Omega)) = \int_{f(\Omega)} \sqrt{|\det G(z)|} \, dz_1 \ldots dz_n,
$$
where $G(z)$ is a metric tensor on $Z$, induced by the scalar product $\langle \cdot \rangle$ on $X$ and the embedding $f$:
$$
G_{ij}(z) = \left\langle \frac{df(z)}{dz^i}, \frac{df(z)}{dz^j} \right\rangle.
$$

Returning to our formula for probability density on $\mathcal{M}$, we now have:
\begin{equation*}
    \begin{split}
        p_\mathcal{M}(f(z)) &=
        p(z) \lim_{\Vol(A) \to 0 \; s.t. \; z \in A} \frac{\int_{A} dz_1 \ldots dz_n}{\int_{f(A)} \sqrt{|\det G(z)|} \, dz_1 \ldots dz_n} \\&=
        p(z) |\det G(z)|^{-1/2}.
    \end{split}
\end{equation*}
Or,
$$
p_\mathcal{M}(x) =
p(f^{-1}(x)) |\det G(f^{-1}(x))|^{-1/2}.
$$

\subsection{Calculation of KL divergence in the case of normalizing flow}

\begin{equation*}
    \begin{split}
        \KL(q_{\MM}(z|x) \| p_{\MM}(z)) &=
        \EE_{z \sim q_{\MM}(z|x)} (\log q_{\MM}(z|x) - \log p_{\MM}(z)) \\&=
        \EE_{z_{hid} \sim q(z_{hid}|x)} (\log q(z_{hid}|x) + \log |\det J_f(z_{hid})|^{-1} \\&- \log p(z_{hid}) - \log |\det J_f(z_{hid})|^{-1}) \\&=
        \EE_{z_{hid} \sim q(z_{hid}|x)} (\log q(z_{hid}|x) - \log p(z_{hid})) \\&=
        \KL(q(z_{hid}|x) \| p(z_{hid})).
    \end{split}
\end{equation*}

where $q(z_{hid}|x)$ is the posterior distribution (i.e. fully-factorized Gauss or Beta) on latent variables of WAE, which we use for manifold embedding, $p(z_{hid})$ is the corresponding prior (i.e. standard Gauss or Uniform), $f$ is the decoder of the WAE, which we use to transform the latent space of WAE into manifold $\MM$, and $J_f(z_{hid})$ is the Jacobian of this transformation. As we see, log-determinants of Jacobians cancel out, and we are left with the KL-divergence on latent space of WAE.

\subsection{Calculation of KL divergence in case of embedding map}

\begin{equation*}
    \begin{split}
        \KL(q_{\MM}(z|x) \| p_{\MM}(z)) &=
        \EE_{z \sim q_{\MM}(z|x)} (\log q_{\MM}(z|x) - \log p_{\MM}(z)) \\&=
        \EE_{z_{hid} \sim q(z_{hid}|x)} (\log q(z_{hid}|x) + \log |\det G(z_{hid})|^{-1/2} \\&- \log p(z_{hid}) - \log |\det G(z_{hid})|^{-1/2}) \\&=
        \EE_{z_{hid} \sim q(z_{hid}|x)} (\log q(z_{hid}|x) - \log p(z_{hid})) \\&=
        \KL(q(z_{hid}|x) \| p(z_{hid})),
    \end{split}
\end{equation*}

where $G$ denotes the metric tensor of the embedding $f$. As in Appendix~4.2, the corresponding terms cancel out.

\end{document}